\documentclass{article}

\usepackage[accepted]{icml2026}

\usepackage{amsmath,amssymb,amsfonts}
\usepackage{graphicx}
\usepackage{booktabs}
\usepackage{hyperref}
\usepackage{url}
\usepackage{algorithm}
\usepackage{algorithmic}
\usepackage{xcolor}
\usepackage{subcaption}
\usepackage{enumitem}
\usepackage{tcolorbox}
\usepackage{soul}
\usepackage{url}

\usepackage{xspace}
\tcbuselibrary{listings, breakable}
\usepackage[capitalize,noabbrev]{cleveref} 

\newcommand{\gemma}{\textsc{Gemma-4-E4B}\xspace}
\newcommand{\apertus}{\textsc{Apertus-8B}\xspace}

\newtcblisting{promptbox}[1]{
    colback=gray!8,
    colframe=gray!40,
    fonttitle=\small\bfseries,
    title=#1,
    listing only,
    listing options={
        basicstyle=\small\ttfamily,
        breaklines=true,
        breakatwhitespace=false,
        columns=fullflexible,
        keepspaces=true,
    },
    left=4pt, right=4pt, top=4pt, bottom=4pt,
    breakable,
}

\begin{document}

\twocolumn[
\icmltitle{Where Do Models Find Happiness? Emotion Vectors in Open-Source LLMs}

\icmlsetsymbol{equal}{*}

  \begin{icmlauthorlist}
    \icmlauthor{Sinie van der Ben}{yyy}
    \icmlauthor{Raphaël Baur}{yyy}
    \icmlauthor{Yannick Metz}{yyy}
    \icmlauthor{Mennatallah El-Assady}{yyy}
  \end{icmlauthorlist}

  \icmlaffiliation{yyy}{Department of Computer Science, ETH, Zurich, Switzerland}
  \icmlcorrespondingauthor{Sinie van der Ben}{vandersi@ethz.ch}

\icmlkeywords{emotion vectors, large language models, replication}

\vskip 0.3in
]

\printAffiliationsAndNotice{}

\begin{abstract}
Recent work identified ``emotion vectors'' in Claude Sonnet 4.5, which are internal representations that encode emotion concepts, causally influence behavior, and exhibit geometry mirroring human psychological structure.
We test the generality of these findings in two open-weight models, \apertus and \gemma, extracting emotion contrast vectors across all layers, using two model-generated corpora. We recover valence geometry for both models, with peak PC1--valence correlations of $r = 0.76$ and $r = 0.83$, approaching the $r = 0.81$ reported for Claude.
Beyond replication, we observe notable differences in how valence representations emerge across model depth. In \gemma, valence is strongly encoded in early layers but collapses towards later layers, whereas \apertus exhibits the opposite pattern, with valence representations absent in early layers, but emerging at mid depths.
Arousal encoding, in contrast, is sensitive to the extraction corpus: both models show stronger PC2--arousal alignment with Gemma-generated stories ($r$ up to $0.45$) than Apertus-generated ones ($r \leq 0.21$), suggesting arousal-relevant cues are unevenly distributed across generated corpora. We open-source our experiment code and dataset for reproducible investigation of emotion representations across language model architectures.
\end{abstract}


\section{Introduction}
As users interact with Large Language Models (LLM), they can encounter responses that appear emotionally reactive, such as expressing frustration when struggling with tasks or enthusiasm when helping users. Recent work by \citet{sofroniew2026twheemotion} moved beyond surface-level observations, identifying internal ``emotion vectors'' in Claude Sonnet 4.5.
They identified 171 linear directions in activation space corresponding to emotion concepts, with correlational and potentially causal relations to model behaviour.
Steering these vectors altered the model's preferences and increased rates of misaligned behaviors such as reward hacking and blackmail. The overall geometry of the emotion space mirrors human psychology, with principal components aligning to valence and arousal axes consistent with Russell's circumplex model \citep{russell1980circumplex}.
These findings raise key questions about generality: \textbf{(1)} Are emotion vectors specific to Claude's training, or a general property of language models' internal representations? \textbf{(2)} How does emotion geometry evolve across layers: Does it emerge suddenly or build up gradually? \textbf{(3)} How does the choice of story corpus affect extraction?
These questions matter for interpretability and safety: If emotion representations are universal and robustly extractable, monitoring them could provide early warnings of misaligned internal states across different models.
We address these questions by replicating and extending emotion vector analysis in two open-weight models: \apertus \cite{swissai2025apertus}, with fully open weights, training data, and code, and \gemma \cite{gemma42026}, a recently released open-source model, both chosen for their relatively small size.
For each model, we extract emotion contrast vectors across multiple layers using two story corpora—one generated by \apertus and one by \gemma—to separate model-intrinsic geometry from corpus-dependent extraction artifacts. Additional related work is provided in \autoref{app:related_work}. We release our code publicly\footnote{\scriptsize\url{https://github.com/sinievanderben/emotion_experiment}}.
\begin{itemize}[itemsep=0em,topsep=0.1em,parsep=0em]
    \item \textbf{Replication of key findings.} We recover valence geometry in both \apertus and \gemma, with the highest PC1--valence correlations of r=0.76 and r=0.83 respectively, demonstrating that emotion vectors generalize beyond Claude to open-weight models across different architectures.
    \item \textbf{Divergent Emergence.} Models differ substantially in when valence structure emerges: \gemma peaks early (layer 16) then fades, while \apertus builds progressively across depth, stabilizing around layer 20. Cross-layer CKA analysis shows a phase transition in \apertus that is absent in Gemma.
    \item \textbf{Corpus-dependent arousal.} Arousal encoding is sensitive to story corpus: both models show substantially stronger PC2--arousal alignment when using Gemma-generated stories (r up to 0.45) than Apertus-generated stories (r $\leq$ 0.17)
\end{itemize}

\section{Methods}
\subsection{Dataset}
We generated \textbf{two} synthetic emotion-story datasets, following \citet{sofroniew2026twheemotion}, with 9 stories for each of 171 emotions. For each emotion, we prompted \apertus and \gemma to write short stories in which characters experience the target emotion without naming it, using a similar prompt to \citet{sofroniew2026twheemotion}. This produced 1{,}539 stories across emotions (Table~\ref{tab:dataset-stats}), plus 40 neutral stories from the same model. The 40 neutral texts form a single fixed set shared by all 171 emotions, since we compute the confound subspace once per layer and project every emotion vector through the same operation. The emotion concepts span the valence-arousal space.

We treat the story corpora as independent variable. By running each model on both Apertus-generated and Gemma-generated stories, we intend to disentangle the emotion findings from corpus-dependent extraction artifacts. No previous work has tested story influence before. 

\subsection{Model}
We analyzed two open-weight language models: \apertus Instruct, a 32-layer transformer, and \gemma, a 42-layer transformer. Both models are instruction-tuned and comparable in scale to enable cross-model comparison of emotion representations. More details on both models can be found in Appendix \ref{app:models}.

\subsection{Contrast Vector Extraction}

Following \citet{sofroniew2026twheemotion}, we construct one activation vector $\mathbf{v}_e^{(l)}$ per emotion $e$ and layer $l$. Since these vectors capture general linguistic structure, we apply a two-step procedure to isolate the emotion-specific component.

First, for each emotion, we perform a forward pass on the corresponding nine stories and cache the residual stream activations at each layer, giving a tensor of shape $(\#\text{tokens}, d_{\text{model}})$ per layer.
Averaging these activations across tokens and stories yields one raw vector $\mathbf{u}_e \in \mathbb{R}^{d_{\text{model}}}$ per emotion and layer, which still mixes emotion-specific and general linguistic features.

Second, we project out non-emotion-specific components. To characterize the emotion-agnostic subspace, we collect mean residual activations from the 40 neutral stories, producing a $(40, d_{\text{model}})$ matrix per layer. PCA on this matrix yields a basis for the subspace; we retain the top $K$
components that together explain 50\% of the variance. To isolate the emotion-specific component, we subtract from each emotion vector its projection onto the neutral subspace to get the contrast vector $\mathbf{v}_e$:
{\setlength{\abovedisplayskip}{3pt}
 \setlength{\belowdisplayskip}{3pt}
 \[
   \mathbf{v}_e = \mathbf{u}_e - \sum_{k=1}^{K} (\mathbf{u}_e \cdot \mathbf{p}_k)\mathbf{p}_k
 \]
}
For \apertus, we extracted vectors from layers \emph{1--31}, and for \gemma, from layers \emph{1--40}. Stacking these vectors across all $|E|$ emotions yields the matrix $V^{(l)} \in \mathbb{R}^{|E| \times d_{\text{model}}}$ at layer $l$, on which we perform the analyses.

\subsection{Analysis}

\noindent\textbf{PCA and Valence-Arousal Correlation}
We applied PCA to the emotion contrast matrix $V^{(l)}$ at each layer and correlated the first two principal components (PC1, PC2) with human valence and arousal ratings from the NRC Valence--Arousal--Dominance Lexicon \cite{mohammad2018obtaining}, following \cite{sofroniew2026twheemotion}. We report Pearson $r$ and corresponding $p$-values.

\noindent\textbf{Cross-layer Representational Similarity with CKA}
We computed linear Centered Kernel Alignment \cite{kornblith2019similarity} between $V^{(l)}$ for all layer pairs within each model and story condition. CKA values near 1 indicate similar representational geometry, while values near 0 indicate orthogonal structure. Because CKA is invariant to orientation in latent space, it is well suited for this comparison and allowed us to quantify how emotion geometry evolves through the network.

\noindent\textbf{Valence direction stability}
Lastly, we identified the valence direction at each layer as the vector most correlated with human valence ratings (using PC1 when this correlation is significant), then computed cosine similarity between these directions across layers to test whether the same subspace encodes valence at different depths.

\section{Results}

\begin{figure*}
    \centering
    \includegraphics[width=\linewidth]{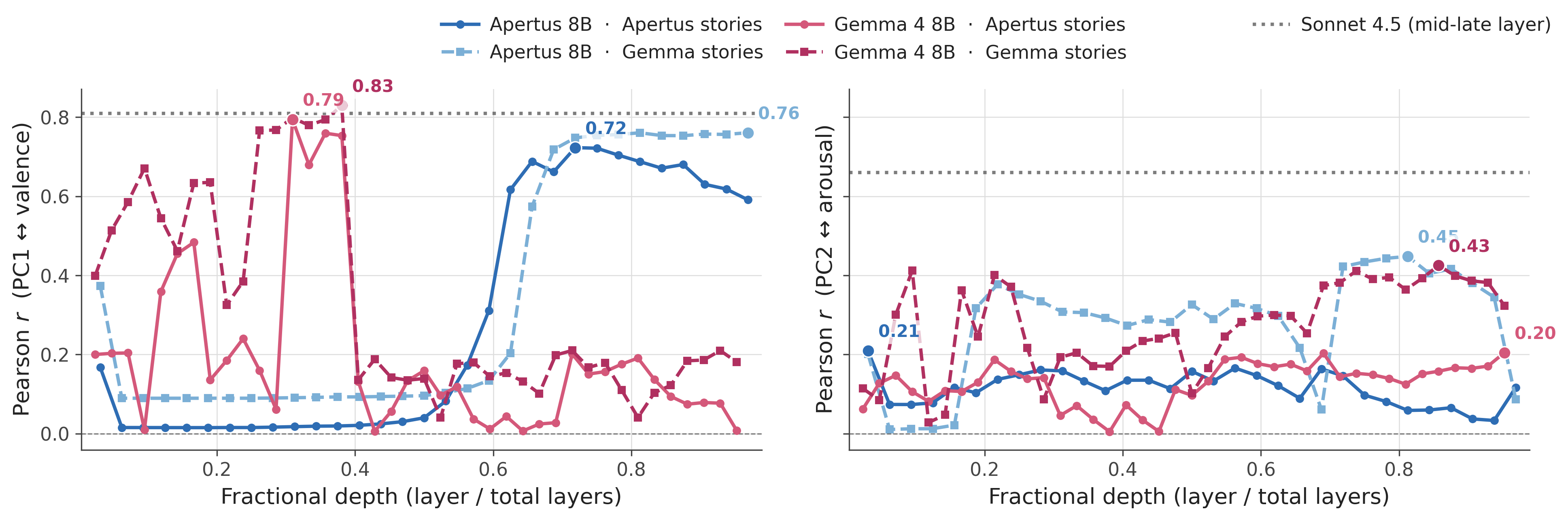}
    \vspace{-0.4em}
    \caption{Pearson correlation between the top two PCs of the
    emotion-vector space and human valence (left, PC1) and arousal (right,
    PC2) across fractional layer depth, for Apertus 8B and \gemma
    probed on Apertus- and Gemma-generated stories (four conditions).
    Hue = model (blue = Apertus, red = Gemma); line style = story source (solid = Apertus, dashed = Gemma). Dotted gray lines mark the Sonnet 4.5 reference at a mid-late layer ($r = 0.81$ valence, $r = 0.66$ arousal; \cite{sofroniew2026twheemotion}).
    }
    \vspace{-0.8em}
    \label{fig:apertus_gemma_valence_arousal_combined}
\end{figure*}

\subsection{Valence Replicates Across Models and Corpora}

The first principal component of the emotion contrast matrix
aligns with human valence ratings in both models, replicating the main result of \cite{sofroniew2026twheemotion}.
\Cref{fig:apertus_gemma_valence_arousal_combined} shows PC1--valence
correlations across fractional layer depth for all four
model$\times$corpus conditions; per-layer values are reported in
\Cref{tab:pc1_valence_per_layer}.
\vspace{-0.5em}
\paragraph{Peak correlations.}
All model$\times$corpus combinations reach a peak between $r = 0.72$ and $0.83$. \apertus peaks at $r = 0.72$ (layer 23, Apertus stories) and $r = 0.76$ (layer 31, Gemma stories); \gemma peaks at $r = 0.79$ (layer 13, Apertus stories) and $r = 0.83$ (layer 16, Gemma stories). All peaks are significant ($p < 10^{-3}$) and approach or exceed the Sonnet 4.5 reference of $r = 0.81$.

\paragraph{Valence Across Network Depth}
Both models reach similar $r$-value peaks with opposite depth profiles. \apertus shows \emph{abrupt late emergence}: PC1–valence correlation is near zero through fractional depth $\approx 0.5$ (layer 17/18), then rises sharply, becoming significant at layer 18 ($r = 0.17$, $p < 0.05$) and exceeding $r = 0.60$ at
layer 21 ($\approx$63\% depth) under both story conditions. 

\gemma instead shows \emph{early encoding followed by collapse}: for Apertus stories, valence peaks at layer 16 ($\approx$38\% depth), then falls near zero by layer 18, with only partial recovery ($r \approx 0.18$--$0.20$) in the final layers. For Gemma stories, the peak comes later and both pre- and post-peak values are higher. The Sonnet 4.5 reference peaks in the mid-late range, indicating that \apertus follows a similar pattern.

\paragraph{Representation space vs.\ valence-axis stability}
To interpret valence trajectories, we examine \textbf{(i)} whole-space representational similarity via linear CKA and \textbf{(ii)} cosine alignment of the layer-wise valence direction for each model–corpus combination. 

\apertus (Figs.~\ref{fig:cka_apertus_apertus_panel},
\ref{fig:cka_apertus_gemma_panel}) shows three CKA phases:
 layers 2--11 form a flat plateau ($\textrm{CKA} \approx 1$); layers 12--21 form a transition band with off-diagonal decay (minimum $0.33$ on Apertus stories, $0.58$ on Gemma stories); and layers 
22--31 form a second plateau. This transition aligns with the rise of PC1--valence correlation, 
suggesting a representational
reorganization. \gemma (Figs.~\ref{fig:cka_gemma_gemma_panel}, \ref{fig:cka_gemma_apertus_panel}) instead shows a smooth gradient across all 40 layers with no sharp transition and CKA $\geq 0.73$ between any pair. The collapse of Gemma's valence correlation around layer 18 therefore cannot stem from a global reorganization, as the geometry remains approximately stable through the collapse.

The valence-direction cosine matrices show what changes. For \apertus on its own stories (Fig.~\ref{fig:cka_apertus_apertus_panel_val}), no off-diagonal cell exceeds $|0.49|$, which can indicate that the recovered direction is noise across layers. On Gemma stories (Fig.~\ref{fig:cka_apertus_gemma_panel_val}), early layers
(2--11) form a coherent block with cosines $0.35$--$0.57$ before becoming noisy in later layers. For \gemma on Gemma stories
(Fig.~\ref{fig:cka_gemma_gemma_panel_val}), there are two positive blocks (layers 2--8 and 9--14) and a late block (28--40), with adjacent-layer cosines up to $\pm 0.55$. On Apertus stories (Fig.~\ref{fig:cka_gemma_apertus_panel_val}) this structure is less pronounced.\\
Because CKA matrices are similar across corpora, the emotion representational space is corpus-invariant. However, the recovered valence axis depends on the input corpus, with Gemma stories yielding cleaner valence directions in both models. Thus, valence is recoverable in both, but not encoded along a consistent axis across depth.

\paragraph{PCA cluster separation at peak layers}
PCA projections at each model's peak layer (Figs.~\ref{fig:apertus_pca}, \ref{fig:gemma_pca}) show emotion clustering and a clear corpus effect. PC1–valence correlations are similar across story conditions (\apertus L23: $0.72$ vs.\ $0.75$; \gemma L13: $0.79$ vs.\ $0.80$), but clusters are more clearly separated for Gemma stories, with positive and negative emotions forming denser groups.

\subsection{Arousal Encoding}

PC2--arousal correlations are generally weaker than PC1--valence and depend strongly on the story corpus
(Fig.~\ref{fig:apertus_gemma_valence_arousal_combined}, right; \Cref{tab:pc2_arousal_per_layer}). On Apertus stories, both models peak below $r = 0.21$ (\apertus: $r = 0.17$ at layer 18; \gemma: $r = 0.21$ at layer 40). On Gemma stories, both models reach $r > 0.40$ (\apertus: $r = 0.45$ at layer 26; \gemma: $r = 0.41$ at layer 31, both $p < 10^{-8}$). 
Possibly, Gemma-generated stories contain more arousal-discriminative linguistic content. We leave a corpus-content analysis to future work.

\section{Discussion}

\noindent\textbf{Main Research Questions} Our results address the three questions raised in the introduction. \textbf{(1)} \emph{Emotion vectors are not specific to Claude's training}. We recover a valence axis of similar
strength in two architecturally distinct open-weight models, with peak correlations matching ($r = 0.83$ for \gemma) or approaching ($r = 0.76$ for
\apertus) the $r = 0.81$ reported for Claude Sonnet 4.5. \textbf{(2)} \emph{Emergence is not uniform across models}. \apertus builds valence alignment abruptly in the second half of the network, while \gemma encodes it early and then loses it mid-network. \textbf{(3)} \emph{The story corpus
 affects extraction}. This is especially clear for arousal: Gemma-generated stories yield correlations more than twice as large as Apertus-generated stories in both probed models.\\
\noindent\textbf{Different paths to the same geometry.}
\gemma and \apertus reach similar peak valence correlations ($r \approx 0.76$--$0.83$) via different layer-wise trajectories. \gemma encodes valence in earlier layers before it degrades in later layers, while \apertus develops it sharply across mid-to-upper layers.
We have not yet explored the possible attribution of this to architecture, training data, or post-training, since the models differ in all three.
Our results show that similar peak valence correlations can hide substantial differences in \textit{where} and \textit{how} valence is computed.\\
\noindent\textbf{Stable representation space, unstable axis}
The representational space (CKA) and valence-axis stability disconnect. In \gemma, the space remains similar across layers even where the PC1–valence correlation collapses, so valence information is preserved. In \apertus, the valence axis is relatively unstable across layers despite a late high plateau of valence–PC1 correlation. Thus, representational similarity between layers does not guarantee a shared valence direction.\\
\noindent\textbf{The arousal gap and corpus dependence}
Arousal shows the weakest replication, but the story-condition analysis suggests that this may be attributed to our methodological choices. With Gemma-generated stories, arousal correlations in both models rise (from $r \leq 0.21$ to $r \geq 0.43$), partially closing the gap with the original result ($r = 0.66$). Because Gemma stories improve arousal extraction in \emph{both} models, the effect likely reflects corpus properties rather than model–story matching. Since it appears in both models, this rules out the simple confound that each model encodes only its own corpus well. We hypothesize that Gemma has the ability to generate stories with greater variation in narrative intensity and physiological arousal cues, so corpus choice for eliciting emotion contrasts is a substantive methodological factor, not an implementation detail. We leave verification to future work.\\

\subsection{Limitations}
Several limitations warrant mention. The first, the original study \cite{sofroniew2026twheemotion} did not release code, so our implementation is reconstructed from the methods they described. Subtle methodological differences may therefore contribute to numerical differences. Second, our analysis covers two open-weight models from two families. Broader cross-architecture comparisons would strengthen claims about how general the valence-pattern is, and whether the trajectory differences generalize to other model families. Third, the corpora we probe are themselves model-generated, which means we cannot fully separate properties of the distributions it produces. A fully model-independent stimulus set would be a stronger control.


\subsection{Future Work}
Several directions follow from our findings and limitations. The most direct is causal validation: steering model outputs at peak-correlation layers along the recovered valence direction would test whether the representational structure we identify is actually used by the model.
Related, the cross-layer rotations of the valence axis raises the question whether steering vectors derived at one layer remain effective when applied to another, even within regions of overall stable space. 
Cross-layer feature tracking using sparse autoencoders could further reveal whether the same interpretable features carry emotion information across the depth ranges we identify, or whether different layers encode emotions through different feature combinations.
Finally, extending this analysis to multi-modal models could test whether the valence axis is preserved across modalities.

\section{Conclusion}
We replicate Anthropic's emotion findings in two open-weight models, achieving valence correlations of $r=0.83$ (\gemma) and $r=0.76$ (\apertus). Cross-layer analysis reveals divergent developmental trajectories: \gemma encodes valence in early layers while \apertus builds it progressively through late layers. These results suggest that similar representations can arise from different computational paths, with implications for layer selection in interpretability work and targeted steering interventions.

\bibliographystyle{icml2026}
\bibliography{references}

@article{cunningham2023sparse,
  title={Sparse Autoencoders Find Highly Interpretable Features in Language Models},
  author={Cunningham, Hoagy and Ewart, Aidan and Riggs, Logan and Huben, Robert and Sharkey, Lee},
  journal={arXiv preprint arXiv:2309.08600},
  year={2023}
}

@article{bricken2023monosemanticity,
  title={Towards Monosemanticity: Decomposing Language Models With Dictionary Learning},
  author={Bricken, Tristan and Templeton, Adam and Batson, Joshua and Chen, Brian and Jermyn, Adam and Conerly, Tom and Turner, Nick and Anil, Cem and Denison, Carson and Askell, Amanda and others},
  journal={Transformer Circuits Thread},
  year={2023}
}

@article{sofroniew2026twheemotion,
  author={Sofroniew, Nicholas and Kauvar, Isaac and Saunders, William and Chen, Runjin and Henighan, Tom and Hydrie, Sasha and Citro, Craig and Pearce, Adam and Tarng, Julius and Gurnee, Wes and Batson, Joshua and Zimmerman, Sam and Rivoire, Kelley and Fish, Kyle and Olah, Chris and Lindsey, Jack},
  title={Emotion Concepts and their Function in a Large Language Model},
  journal={Transformer Circuits Thread},
  year={2026},
  url={https://transformer-circuits.pub/2026/emotions/index.html}
}

@misc{swissai2025apertus,
  title={{Apertus: Democratizing Open and Compliant LLMs for Global Language Environments}},
  author={Alejandro Hernández-Cano and Alexander Hägele and Allen Hao Huang and Angelika Romanou and Antoni-Joan Solergibert and Barna Pasztor and Bettina Messmer and Dhia Garbaya and Eduard Frank Ďurech and Ido Hakimi and Juan García Giraldo and Mete Ismayilzada and Negar Foroutan and Skander Moalla and Tiancheng Chen and Vinko Sabolčec and Yixuan Xu and Michael Aerni and Badr AlKhamissi and Ines Altemir Marinas and Mohammad Hossein Amani and Matin Ansaripour and Ilia Badanin and Harold Benoit and Emanuela Boros and Nicholas Browning and Fabian Bösch and Maximilian Böther and Niklas Canova and Camille Challier and Clement Charmillot and Jonathan Coles and Jan Deriu and Arnout Devos and Lukas Drescher and Daniil Dzenhaliou and Maud Ehrmann and Dongyang Fan and Simin Fan and Silin Gao and Miguel Gila and María Grandury and Diba Hashemi and Alexander Hoyle and Jiaming Jiang and Mark Klein and Andrei Kucharavy and Anastasiia Kucherenko and Frederike Lübeck and Roman Machacek and Theofilos Manitaras and Andreas Marfurt and Kyle Matoba and Simon Matrenok and Henrique Mendoncça and Fawzi Roberto Mohamed and Syrielle Montariol and Luca Mouchel and Sven Najem-Meyer and Jingwei Ni and Gennaro Oliva and Matteo Pagliardini and Elia Palme and Andrei Panferov and Léo Paoletti and Marco Passerini and Ivan Pavlov and Auguste Poiroux and Kaustubh Ponkshe and Nathan Ranchin and Javi Rando and Mathieu Sauser and Jakhongir Saydaliev and Muhammad Ali Sayfiddinov and Marian Schneider and Stefano Schuppli and Marco Scialanga and Andrei Semenov and Kumar Shridhar and Raghav Singhal and Anna Sotnikova and Alexander Sternfeld and Ayush Kumar Tarun and Paul Teiletche and Jannis Vamvas and Xiaozhe Yao and Hao Zhao Alexander Ilic and Ana Klimovic and Andreas Krause and Caglar Gulcehre and David Rosenthal and Elliott Ash and Florian Tramèr and Joost VandeVondele and Livio Veraldi and Martin Rajman and Thomas Schulthess and Torsten Hoefler and Antoine Bosselut and Martin Jaggi and Imanol Schlag},
  year={2025},
  howpublished={\url{https://arxiv.org/abs/2509.14233}}
}

@inproceedings{
tigges2024language,
title={Language Models Linearly Represent Sentiment},
author={Curt Tigges and Oskar John Hollinsworth and Atticus Geiger and Neel Nanda},
booktitle={ICML 2024 Workshop on Mechanistic Interpretability},
year={2024},
url={https://openreview.net/forum?id=Xsf6dOOMMc}
}

@inproceedings{mikolov2013linguistic,
    title = "Linguistic Regularities in Continuous Space Word Representations",
    author = "Mikolov, Tomas  and
      Yih, Wen-tau  and
      Zweig, Geoffrey",
    editor = "Vanderwende, Lucy  and
      Daum{\'e} III, Hal  and
      Kirchhoff, Katrin",
    booktitle = "Proceedings of the 2013 Conference of the North {A}merican Chapter of the Association for Computational Linguistics: Human Language Technologies",
    month = jun,
    year = "2013",
    address = "Atlanta, Georgia",
    publisher = "Association for Computational Linguistics",
    url = "https://aclanthology.org/N13-1090/",
    pages = "746--751"
}

@inproceedings{
park2023linear,
title={The Linear Representation Hypothesis and the Geometry of Large Language Models},
author={Kiho Park and Yo Joong Choe and Victor Veitch},
booktitle={Causal Representation Learning Workshop at NeurIPS 2023},
year={2023},
url={https://openreview.net/forum?id=T0PoOJg8cK}
}

@article{elhage2022superposition,
   title={Toy Models of Superposition},
   author={Elhage, Nelson and Hume, Tristan and Olsson, Catherine and Schiefer, Nicholas and Henighan, Tom and Kravec, Shauna and Hatfield-Dodds, Zac and Lasenby, Robert and Drain, Dawn and Chen, Carol and Grosse, Roger and McCandlish, Sam and Kaplan, Jared and Amodei, Dario and Wattenberg, Martin and Olah, Christopher},
   year={2022},
   journal={Transformer Circuits Thread},
   url={https://transformer-circuits.pub/2022/toy_model/index.html}
}

@misc{marks2024geometry,
      title={The Geometry of Truth: Emergent Linear Structure in Large Language Model Representations of True/False Datasets}, 
      author={Samuel Marks and Max Tegmark},
      year={2024},
      eprint={2310.06824},
      archivePrefix={arXiv},
      primaryClass={cs.AI},
      url={https://arxiv.org/abs/2310.06824}, 
}

@inproceedings{
arditi2024refusal,
title={Refusal in Language Models Is Mediated by a Single Direction},
author={Andy Arditi and Oscar Balcells Obeso and Aaquib Syed and Daniel Paleka and Nina Rimsky and Wes Gurnee and Neel Nanda},
booktitle={The Thirty-eighth Annual Conference on Neural Information Processing Systems},
year={2024},
url={https://openreview.net/forum?id=pH3XAQME6c}
}

@inproceedings{rimsky2024steering,
    title = "Steering Llama 2 via Contrastive Activation Addition",
    author = "Rimsky, Nina  and
      Gabrieli, Nick  and
      Schulz, Julian  and
      Tong, Meg  and
      Hubinger, Evan  and
      Turner, Alexander",
    editor = "Ku, Lun-Wei  and
      Martins, Andre  and
      Srikumar, Vivek",
    booktitle = "Proceedings of the 62nd Annual Meeting of the Association for Computational Linguistics (Volume 1: Long Papers)",
    month = aug,
    year = "2024",
    address = "Bangkok, Thailand",
    publisher = "Association for Computational Linguistics",
    url = "https://aclanthology.org/2024.acl-long.828/",
    doi = "10.18653/v1/2024.acl-long.828",
    pages = "15504--15522",
}

@misc{radford2017sentiment,
      title={Learning to Generate Reviews and Discovering Sentiment}, 
      author={Alec Radford and Rafal Jozefowicz and Ilya Sutskever},
      year={2017},
      eprint={1704.01444},
      archivePrefix={arXiv},
      primaryClass={cs.LG},
      url={https://arxiv.org/abs/1704.01444}, 
}

@inproceedings{
valeriani2023geometry,
title={The geometry of hidden representations of large transformer models},
author={Lucrezia Valeriani and Diego Doimo and Francesca Cuturello and Alessandro Laio and Alessio ansuini and Alberto Cazzaniga},
booktitle={Thirty-seventh Conference on Neural Information Processing Systems},
year={2023},
url={https://openreview.net/forum?id=cCYvakU5Ek}
}

@inproceedings{
cheng2025emergence,
title={Emergence of a High-Dimensional Abstraction Phase in Language Transformers},
author={Emily Cheng and Diego Doimo and Corentin Kervadec and Iuri Macocco and Lei Yu and Alessandro Laio and Marco Baroni},
booktitle={The Thirteenth International Conference on Learning Representations},
year={2025},
url={https://openreview.net/forum?id=0fD3iIBhlV}
}

@inproceedings{mohammad2018obtaining,
  title={Obtaining Reliable Human Ratings of Valence, Arousal, and Dominance for 20,000 English Words},
  author={Mohammad, Saif M.},
  booktitle={Proceedings of ACL},
  year={2018}
}

@article{russell1980circumplex,
  title={A circumplex model of affect.},
  author={Russell, James A},
  journal={Journal of personality and social psychology},
  volume={39},
  number={6},
  pages={1161},
  year={1980},
  publisher={American Psychological Association}
}

@inproceedings{kornblith2019similarity,
  title={Similarity of neural network representations revisited},
  author={Kornblith, Simon and Norouzi, Mohammad and Lee, Honglak and Hinton, Geoffrey},
  booktitle={International conference on machine learning},
  pages={3519--3529},
  year={2019},
  organization={PMlR}
}

@misc{gemma42026,
  title = {Gemma 4: Expanding the Gemmaverse with Apache 2.0},
  author = {Google DeepMind},
  year = {2026},
  url = {https://opensource.googleblog.com/2026/03/gemma-4-expanding-the-gemmaverse-with-apache-20.html},
  note = {Accessed: 2026-04-28}
}

@misc{choi2026latent,
      title={Latent Structure of Affective Representations in Large Language Models}, 
      author={Benjamin J. Choi and Melanie Weber},
      year={2026},
      eprint={2604.07382},
      archivePrefix={arXiv},
      primaryClass={cs.LG},
      url={https://arxiv.org/abs/2604.07382}, 
}

@misc{sun2026valence,
      title={Valence-Arousal Subspace in LLMs: Circular Emotion Geometry and Multi-Behavioral Control}, 
      author={Lihao Sun and Lewen Yan and Xiaoya Lu and Andrew Lee and Jie Zhang and Jing Shao},
      year={2026},
      eprint={2604.03147},
      archivePrefix={arXiv},
      primaryClass={cs.CL},
      url={https://arxiv.org/abs/2604.03147}, 
}

\appendix

\onecolumn



\section{Related Work}
\label{app:related_work}
\noindent\textbf{Linear representations in LLMs.} The linear representation hypothesis holds that high-level concepts are encoded as directions in activation space \citep{mikolov2013linguistic, elhage2022superposition, park2023linear}. \citet{tigges2024language} demonstrated this for sentiment, finding a single direction captures positive-negative valence across tasks.
Subsequent work extended linear representations to truth \citep{marks2024geometry}, refusal \citep{arditi2024refusal}, and behavioral tendencies \citep{rimsky2024steering}.
Sparse autoencoders can extract directions at scale, decomposing polysemantic activations into interpretable features \citep{bricken2023monosemanticity, cunningham2023sparse}.\\
\noindent\textbf{Emotion in language models.} Early work identified a ``sentiment neuron'' in LSTMs \cite{radford2017sentiment}, though later analysis suggested emotional content is distributed across many neurons. \cite{sofroniew2026twheemotion} provide a comprehensive analysis, extracting 171 emotion vectors from Claude Sonnet 4.5 and demonstrating causal influence on behavior.
They found emotion geometry mirrors human psychological structure, with valence and arousal as principal axes.  Concurrent work extends this to other models: \cite{sun2026valence} identify a valence-arousal subspace in Llama and Qwen with circumplex-consistent circular geometry, where steering along VA axes controls refusal and sycophancy. 
\cite{choi2026latent} find coherent affective representations in Gemma-2, Mistral, and LLaMA with modest nonlinear global structure. 
We build on \citet{sofroniew2026twheemotion}, testing generalization across architectures and the role of extraction methodology.\\
\noindent\textbf{Cross-layer geometry.} Transformer representations evolve across layers in characteristic ways. \citet{valeriani2023geometry} found intrinsic dimension expands then contracts, with semantics concentrated at intermediate depths. \cite{cheng2025emergence} identified a "high-dimensional abstraction phase" where representations peak in complexity before simplifying toward outputs.

\newpage

\section{Story Dataset}
The emotion story datasets was generated using \apertus and \gemma, following a methodology similar to Anthropic's emotion vectors work \cite{sofroniew2026twheemotion}. The 171 emotions were copied from their work. Stories were designed to convey emotions implicitly, such as never naming the target emotion directly, but instead relying instead on character actions, physical sensations, dialogue, and situational context. The prompts used were also similar, to introduce as little methodological confound as possible. 

\subsection{Dataset statistics}

\begin{table}[h]
\centering
\caption{Emotion story dataset statistics by corpus. 
Apertus stories were deduplicated to match 
the uniform 9-stories-per-topic structure of the Gemma corpus.}
\label{tab:dataset-stats}
\begin{tabular}{lll}
\toprule
\textbf{Statistic} & \textbf{Apertus stories} & \textbf{Gemma stories} \\
\midrule
Generator model & \apertus-Instruct-2509 & \gemma-it \\
Total emotions & 171 & 171 \\
Unique topics & 100 & 100 \\
Stories per emotion & 9 & 9 \\
Total stories & 1,539 & 1,539 \\
Mean story length & $\sim$215 words & $\sim$144 words \\
Story length range & 65--665 words & 81--298 words \\
\bottomrule
\end{tabular}
\end{table}

\subsection{Activation collection}
\label{app:models}
Residual stream activations were collected from \apertus-Instruct at multiple transformer layers (Table \ref{tab:activation-config}). The model can be found through HuggingFace: \texttt{swiss-ai/\apertus-Instruct-2509}.

For Gemma, different layers were picked to collect activations from (Table \ref{tab:activation-config}). The model was also accessed through HuggingFace: 
\texttt{google/\gemma-it}.

\begin{table}[h]
\centering
\caption{Activation extraction configuration for both models.}
\label{tab:activation-config}
\begin{tabular}{lll}
\toprule
\textbf{Parameter} & \textbf{\apertus} & \textbf{\gemma 8B} \\
\midrule
Model & \apertus-Instruct-2509 & \gemma \\
Layers collected & 1-31 & 1-40 \\
Hook location & \texttt{resid\_post} & \texttt{resid\_post} \\
Batch size & 8 sequences & 8 sequences \\
Max sequence length & 1024 tokens & 1024 tokens \\
Dataset & Pile (uncopyrighted), train split & Pile (uncopyrighted), train split \\
\bottomrule
\end{tabular}
\end{table}

\section{Additional Results}
\label{appendix:full_tables}

\subsection{Principal Component Valence}
\begin{table}[H]
  \centering
\small
\caption{PC1--valence (Pearson $r$) across layers and story conditions. Bold indicates the peak layer per model--condition pair. $^{\dagger} p<0.05$; $^{\ddagger} p<0.01$; $^{*} p<0.001$.}
\label{tab:pc1_valence_per_layer}
\begin{tabular}{cc cc cc cc}
\toprule
& & \multicolumn{2}{c}{\textbf{\apertus}} & & & \multicolumn{2}{c}{\textbf{\gemma 8B}} \\
\cmidrule{3-4} \cmidrule{7-8}
& & \textit{\apertus stories} & \textit{\gemma stories} & & & \textit{\apertus stories} & \textit{\gemma stories} \\
\textbf{Layer} & & $r$ & $r$ & \textbf{Layer} & & $r$ & $r$ \\
\midrule
    1  & & $0.1675^{\dagger}$        & $0.3737^{*}$              & 1  & & $0.2000^{\ddagger}$       & $0.3986^{*}$              \\
    2  & & 0.0154                    & 0.0895                    & 2  & & $0.2030^{\ddagger}$       & $0.5138^{*}$              \\
    3  & & 0.0152                    & 0.0894                    & 3  & & $0.2040^{\ddagger}$       & $0.5857^{*}$              \\
    4  & & 0.0152                    & 0.0894                    & 4  & & 0.0098                    & $0.6708^{*}$              \\
    5  & & 0.0151                    & 0.0894                    & 5  & & $0.3592^{*}$              & $0.5443^{*}$              \\
    6  & & 0.0151                    & 0.0895                    & 6  & & $0.4550^{*}$              & $0.4619^{*}$              \\
    7  & & 0.0154                    & 0.0895                    & 7  & & $0.4835^{*}$              & $0.6333^{*}$              \\
    8  & & 0.0153                    & 0.0894                    & 8  & & 0.1359                    & $0.6352^{*}$              \\
    9  & & 0.0163                    & 0.0900                    & 9  & & $0.1851^{\dagger}$        & $0.3248^{*}$              \\
    10 & & 0.0176                    & 0.0908                    & 10 & & $0.2401^{\ddagger}$       & $0.3851^{*}$              \\
    11 & & 0.0188                    & 0.0916                    & 11 & & $0.1598^{\dagger}$        & $0.7664^{*}$              \\
    12 & & 0.0193                    & 0.0928                    & 12 & & 0.0614                    & $0.7675^{*}$              \\
    13 & & 0.0209                    & 0.0928                    & 13 & & $\mathbf{0.7940}^{*}$     & $0.7984^{*}$              \\
    14 & & 0.0246                    & 0.0938                    & 14 & & $0.6797^{*}$              & $0.7795^{*}$              \\
    15 & & 0.0302                    & 0.0946                    & 15 & & $0.7595^{*}$              & $0.7938^{*}$              \\
    16 & & 0.0400                    & 0.0959                    & 16 & & $0.7533^{*}$              & $\mathbf{0.8296}^{*}$     \\
    17 & & 0.0821                    & 0.1034                    & 17 & & 0.1318                    & 0.1360                    \\
    18 & & $0.1732^{\dagger}$        & 0.1141                    & 18 & & 0.0060                    & $0.1880^{\dagger}$        \\
    19 & & $0.3112^{*}$              & 0.1343                    & 19 & & 0.0564                    & 0.1418                    \\
    20 & & $0.6169^{*}$              & $0.2034^{\ddagger}$       & 20 & & 0.1337                    & 0.1350                    \\
    21 & & $0.6882^{*}$              & $0.5739^{*}$              & 21 & & $0.1591^{\dagger}$        & 0.1393                    \\
    22 & & $0.6616^{*}$              & $0.7181^{*}$              & 22 & & 0.0968                    & 0.0403                    \\
    23 & & $\mathbf{0.7230}^{*}$     & $0.7478^{*}$              & 23 & & 0.1187                    & $0.1769^{\dagger}$        \\
    24 & & $0.7216^{*}$              & $0.7539^{*}$              & 24 & & 0.0370                    & $0.1796^{\dagger}$        \\
    25 & & $0.7040^{*}$              & $0.7561^{*}$              & 25 & & 0.0117                    & 0.1446                    \\
    26 & & $0.6875^{*}$              & ${0.7606}^{*}$            & 26 & & 0.0436                    & $0.1536^{\dagger}$        \\
    27 & & $0.6710^{*}$              & $0.7535^{*}$              & 27 & & 0.0067                    & 0.1319                    \\
    28 & & $0.6803^{*}$              & $0.7533^{*}$              & 28 & & 0.0242                    & 0.1015                    \\
    29 & & $0.6302^{*}$              & $0.7573^{*}$              & 29 & & 0.0277                    & $0.1986^{\ddagger}$       \\
    30 & & $0.6183^{*}$              & $0.7564^{*}$              & 30 & & $0.2014^{\ddagger}$       & $0.2101^{\ddagger}$       \\
    31 & & $0.5910^{*}$              & $\mathbf{0.7608}^{*}$     & 31 & & $0.1505^{\dagger}$        & $0.1672^{\dagger}$        \\
    32 & & ---                       & ---                       & 32 & & $0.1567^{\dagger}$        & $0.1793^{\dagger}$        \\
    33 & & ---                       & ---                       & 33 & & $0.1754^{\dagger}$        & 0.1105                    \\
    34 & & ---                       & ---                       & 34 & & $0.1910^{\dagger}$        & 0.0408                    \\
    35 & & ---                       & ---                       & 35 & & 0.1366                    & 0.1034                    \\
    36 & & ---                       & ---                       & 36 & & 0.0939                    & 0.1222                    \\
    37 & & ---                       & ---                       & 37 & & 0.0743                    & $0.1841^{\dagger}$        \\
    38 & & ---                       & ---                       & 38 & & 0.0786                    & $0.1861^{\dagger}$        \\
    39 & & ---                       & ---                       & 39 & & 0.0769                    & $0.2099^{\ddagger}$       \\
    40 & & ---                       & ---                       & 40 & & 0.0081                    & $0.1810^{\dagger}$        \\
\bottomrule
\end{tabular}
\end{table}

\subsection{Principal Component Arousal}

\begin{table}[H]
  \centering
\small
\caption{PC2--arousal (Pearson $r$) across layers and story conditions. Bold indicates the peak layer per model--condition pair. $^{\dagger} p<0.05$; $^{\ddagger} p<0.01$; $^{*} p<0.001$.}
\label{tab:pc2_arousal_per_layer}
\begin{tabular}{cc cc cc cc}
\toprule
& & \multicolumn{2}{c}{\textbf{\apertus}} & & & \multicolumn{2}{c}{\textbf{\gemma}} \\
\cmidrule{3-4} \cmidrule{7-8}
& & \textit{\apertus stories} & \textit{\gemma stories} & & & \textit{\apertus stories} & \textit{\gemma stories} \\
\textbf{Layer} & & $r$ & $r$ & \textbf{Layer} & & $r$ & $r$ \\
\midrule
    1  & & $\mathbf{0.2093}^{\ddagger}$       & $0.2012^{\ddagger}$       & 1  & & 0.0624                    & 0.1141                    \\
    2  & & 0.0737                    & 0.0105                    & 2  & & 0.1273                    & 0.0845                    \\
    3  & & 0.0735                    & 0.0122                    & 3  & & $0.1472^{\dagger}$        & $0.3005^{*}$              \\
    4  & & 0.0778                    & 0.0124                    & 4  & & 0.1060                    & $0.4120^{*}$              \\
    5  & & 0.1168                    & 0.0214                    & 5  & & 0.0815                    & 0.0282                    \\
    6  & & 0.1032                    & $0.3171^{*}$              & 6  & & 0.1086                    & 0.0476                    \\
    7  & & 0.1370                    & $0.3772^{*}$              & 7  & & 0.1064                    & $0.3625^{*}$              \\
    8  & & $0.1488^{\dagger}$        & $0.3520^{*}$              & 8  & & 0.1298                    & $0.2450^{\ddagger}$       \\
    9  & & $0.1610^{\dagger}$        & $0.3345^{*}$              & 9  & & $0.1867^{\dagger}$        & $0.4008^{*}$              \\
    10 & & $0.1585^{\dagger}$        & $0.3076^{*}$              & 10 & & $0.1574^{\dagger}$        & $0.3710^{*}$              \\
    11 & & 0.1328                    & $0.3060^{*}$              & 11 & & 0.1384                    & $0.2168^{\ddagger}$       \\
    12 & & 0.1085                    & $0.2923^{*}$              & 12 & & 0.1414                    & 0.0872                    \\
    13 & & 0.1350                    & $0.2731^{*}$              & 13 & & 0.0459                    & $0.1928^{\dagger}$        \\
    14 & & 0.1351                    & $0.2883^{*}$              & 14 & & 0.0707                    & $0.2045^{\ddagger}$       \\
    15 & & 0.1137                    & $0.2818^{*}$              & 15 & & 0.0360                    & $0.1709^{\dagger}$        \\
    16 & & $0.1573^{\dagger}$        & $0.3262^{*}$              & 16 & & 0.0048                    & $0.1700^{\dagger}$        \\
    17 & & 0.1329                    & $0.2890^{*}$              & 17 & & 0.0726                    & $0.2094^{\ddagger}$       \\
    18 & & $0.1657^{\dagger}$        & $0.3290^{*}$              & 18 & & 0.0345                    & $0.2341^{\ddagger}$       \\
    19 & & $0.1469^{\dagger}$        & $0.3166^{*}$              & 19 & & 0.0054                    & $0.2407^{\ddagger}$       \\
    20 & & 0.1211                    & $0.2974^{*}$              & 20 & & 0.1114                    & $0.2547^{*}$              \\
    21 & & 0.0891                    & $0.2167^{\ddagger}$       & 21 & & 0.0965                    & 0.1018                    \\
    22 & & $0.1637^{\dagger}$        & 0.0607                    & 22 & & 0.1325                    & $0.1660^{\dagger}$        \\
    23 & & $0.1459^{\dagger}$        & $0.4222^{*}$              & 23 & & $0.1879^{\dagger}$        & $0.2452^{\ddagger}$       \\
    24 & & 0.0972                    & $0.4341^{*}$              & 24 & & $0.1931^{\dagger}$        & $0.2819^{*}$              \\
    25 & & 0.0810                    & $0.4432^{*}$              & 25 & & $0.1769^{\dagger}$        & $0.2970^{*}$              \\
    26 & & 0.0586                    & $\mathbf{0.4480}^{*}$     & 26 & & $0.1690^{\dagger}$        & $0.2997^{*}$              \\
    27 & & 0.0597                    & $0.4049^{*}$              & 27 & & $0.1755^{\dagger}$        & $0.2979^{*}$              \\
    28 & & 0.0654                    & $0.4165^{*}$              & 28 & & $0.1585^{\dagger}$        & $0.2538^{*}$              \\
    29 & & 0.0371                    & $0.3804^{*}$              & 29 & & $0.2030^{\ddagger}$       & $0.3741^{*}$              \\
    30 & & 0.0331                    & $0.3449^{*}$              & 30 & & 0.1437                    & $0.3816^{*}$              \\
    31 & & 0.1165                    & 0.0866                    & 31 & & $0.1524^{\dagger}$        & $0.4115^{*}$              \\
    32 & & ---                       & ---                       & 32 & & 0.1487                    & $0.3904^{*}$              \\
    33 & & ---                       & ---                       & 33 & & 0.1393                    & $0.3950^{*}$              \\
    34 & & ---                       & ---                       & 34 & & 0.1248                    & $0.3638^{*}$              \\
    35 & & ---                       & ---                       & 35 & & $0.1509^{\dagger}$        & $0.3926^{*}$              \\
    36 & & ---                       & ---                       & 36 & & $0.1577^{\dagger}$        & $\mathbf{0.4251}^{*}$     \\
    37 & & ---                       & ---                       & 37 & & $0.1665^{\dagger}$        & $0.3989^{*}$              \\
    38 & & ---                       & ---                       & 38 & & $0.1650^{\dagger}$        & $0.3863^{*}$              \\
    39 & & ---                       & ---                       & 39 & & $0.1706^{\dagger}$        & $0.3818^{*}$              \\
    40 & & ---                       & ---                       & 40 & & $\mathbf{0.2047}^{\ddagger}$ & $0.3234^{*}$           \\
\bottomrule
\end{tabular}
\end{table}

\newpage
\subsection{CKA Figures}
CKA is a measure of how similar the emotion space is between 2 layers. The diagonal is always 1, which is a layer compared to itself. 

\begin{itemize}
    \item CKA close to 1: spatial arrangement of emotion vectors between 2 layers is nearly identical. 
    \item CKA close to 0: spatial arrangement has changed substantially between 2 layers
\end{itemize}

So each cell answers: does the model organize emotions in the same way at layer A as in layer B? The higher the value, the more similar. 

\subsubsection{\textbf{\apertus} CKA results}

\begin{figure}[H]
\caption{\textbf{\apertus} CKA results on Apertus stories}
\includegraphics[width=1.0\textwidth]{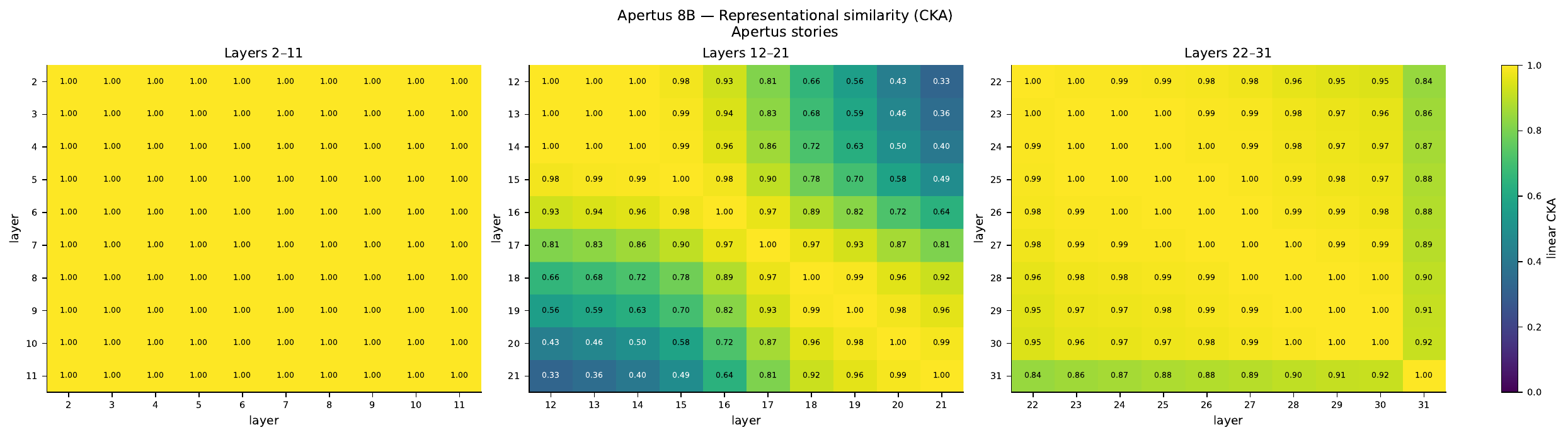}
\centering
 \label{fig:cka_apertus_apertus_panel}
\end{figure}

\begin{figure}[H]
\caption{\textbf{\apertus} CKA values on Gemma stories }
\includegraphics[width=1.0\textwidth]{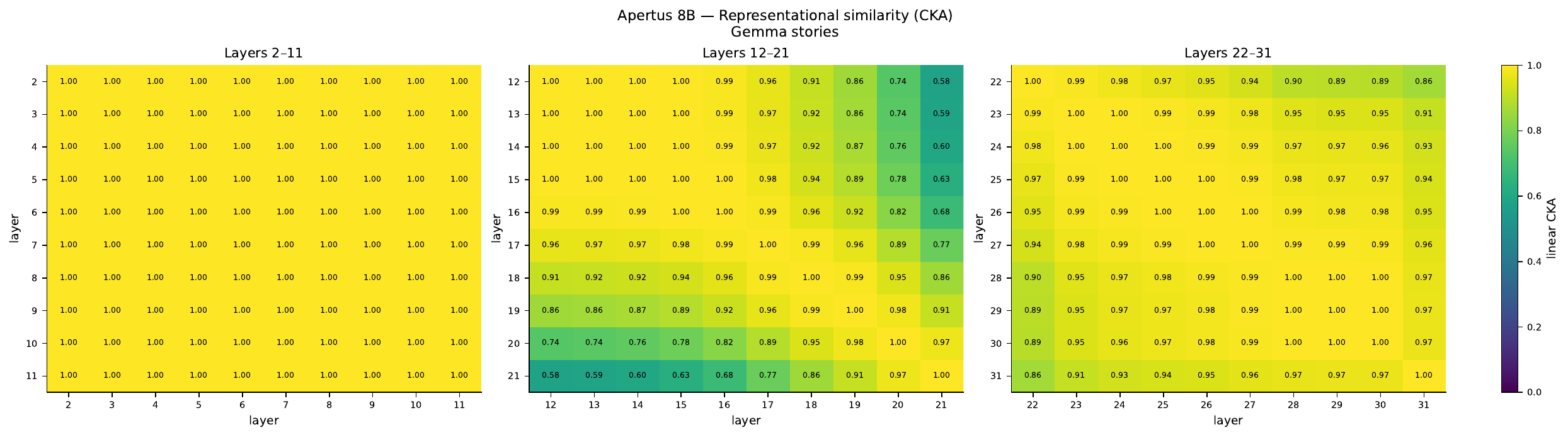}
\centering
 \label{fig:cka_apertus_gemma_panel}
\end{figure}

\subsubsection{\gemma CKA results}

\begin{figure}[H]
\caption{\textbf{\gemma} CKA values on Gemma stories}
\includegraphics[width=1.0\textwidth]{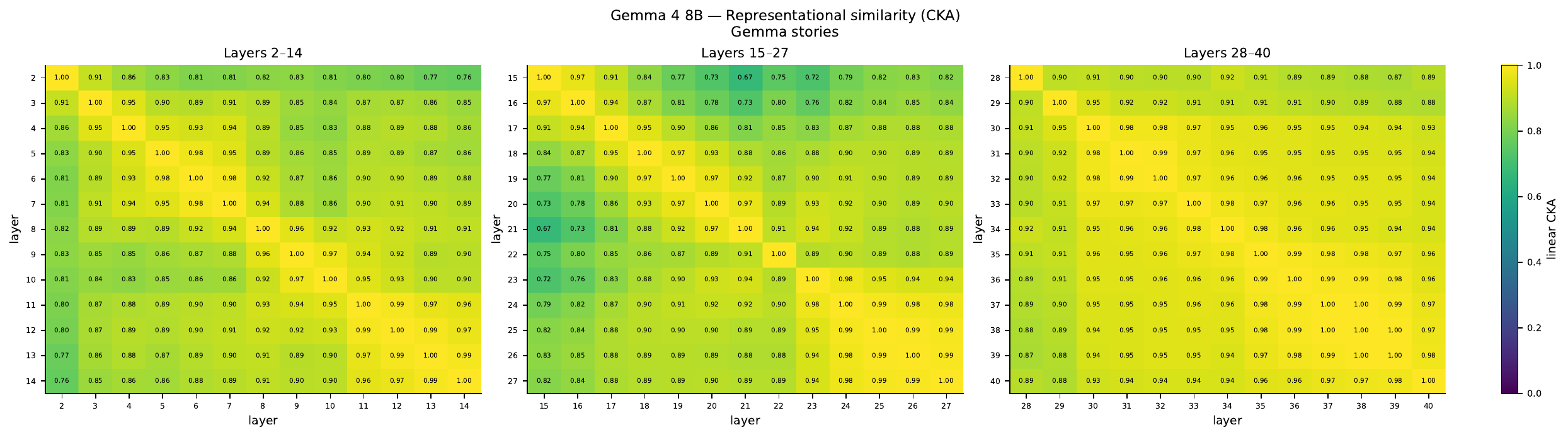}
\centering
 \label{fig:cka_gemma_gemma_panel}
\end{figure}

\begin{figure}[H]
\caption{\textbf{\gemma} CKA values on Apertus stories }
\includegraphics[width=1.0\textwidth]{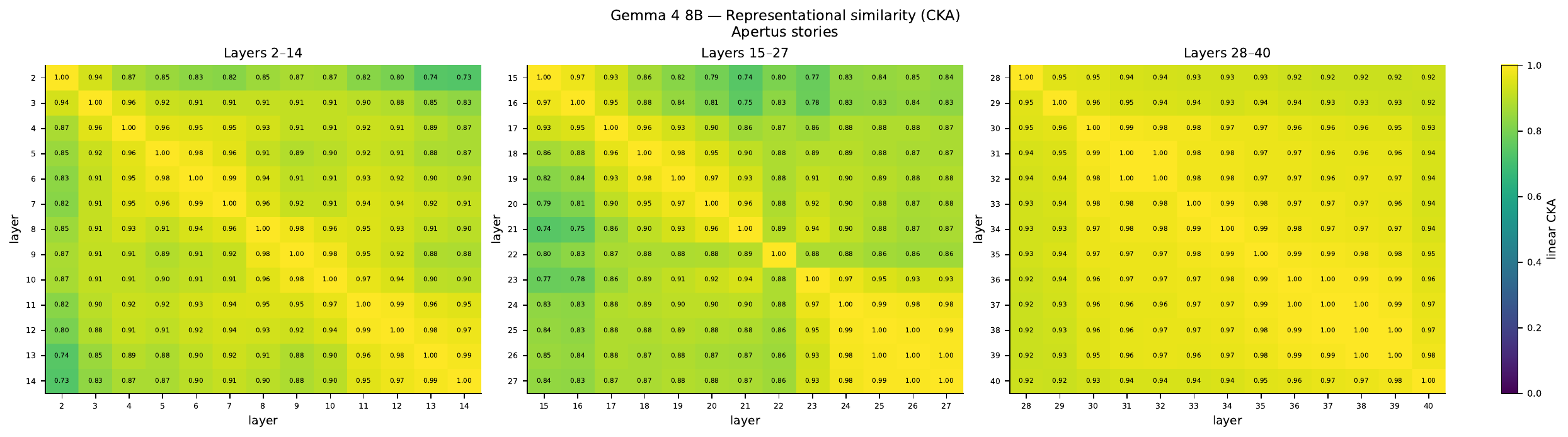}
\centering
 \label{fig:cka_gemma_apertus_panel}
\end{figure}

\subsection{Valence Direction Alignment}
Each cell shows the cosine similarity between the valence direction vectors at 2 layers. The valence direction is the axis in activation space that best predicts the emotion valence. 

\begin{itemize}
    \item \textbf{Cosine similarity close to 1}. Valence axis points in the same direction in both layers, consistent positive axis.
    \item \textbf{Cosine similarity close to 0}. Valence axes are orthogonal, they've rotated completely.
    \item \textbf{Cosine similarity close to -1}. The axis has flipped direction.
\end{itemize}

CKA provides information about the whole space, while the cosine similarity specifically shows whether the valence axis is stable. A predominantly blue matrix would indicate that the model has a persistent stable direction to represent positive vs. negative emotions across many layers. 

The valence direction stability line plot shows the cosine similarity between 2 adjacent layers. It has a similar interpretation as the values in the panel, but only for adjacent layers. The interpretation can be slightly different, because it shows if the valence axis points in the same direction from layer to layer. A dip reveals a specific transition, where the model changes how it encodes valence. 

\subsubsection{\textbf{\apertus} Valence Alignment}

\begin{figure}[H]
\caption{\textbf{\apertus} validation on Apertus stories }
\includegraphics[width=1.0\textwidth]{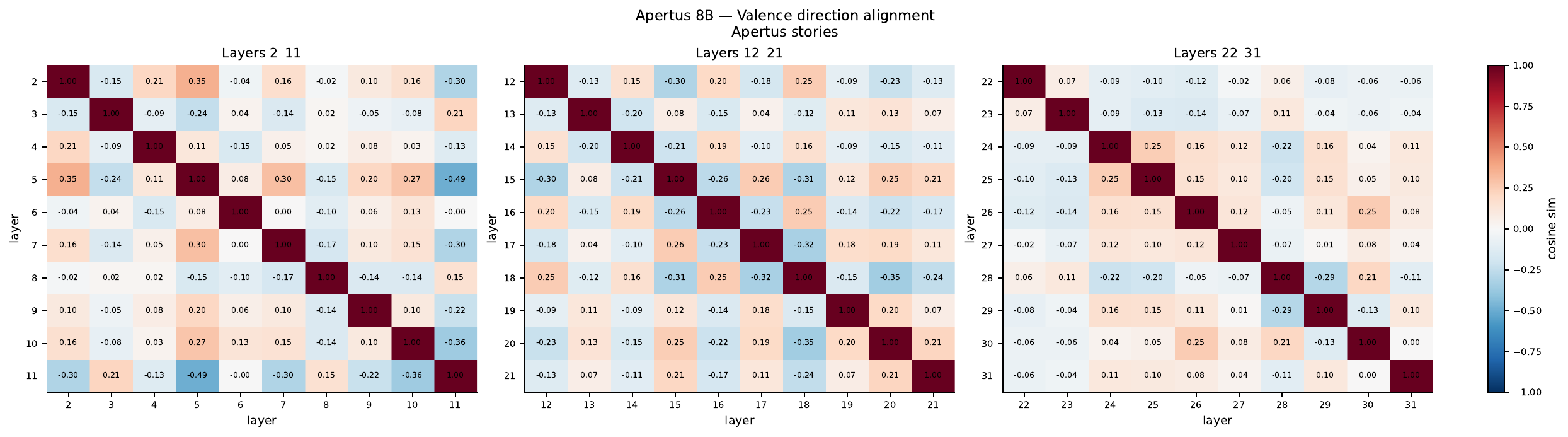}
\centering
 \label{fig:cka_apertus_apertus_panel_val}
\end{figure}

\begin{figure}[H]
\caption{\textbf{\apertus} validation on Gemma stories }
\includegraphics[width=1.0\textwidth]{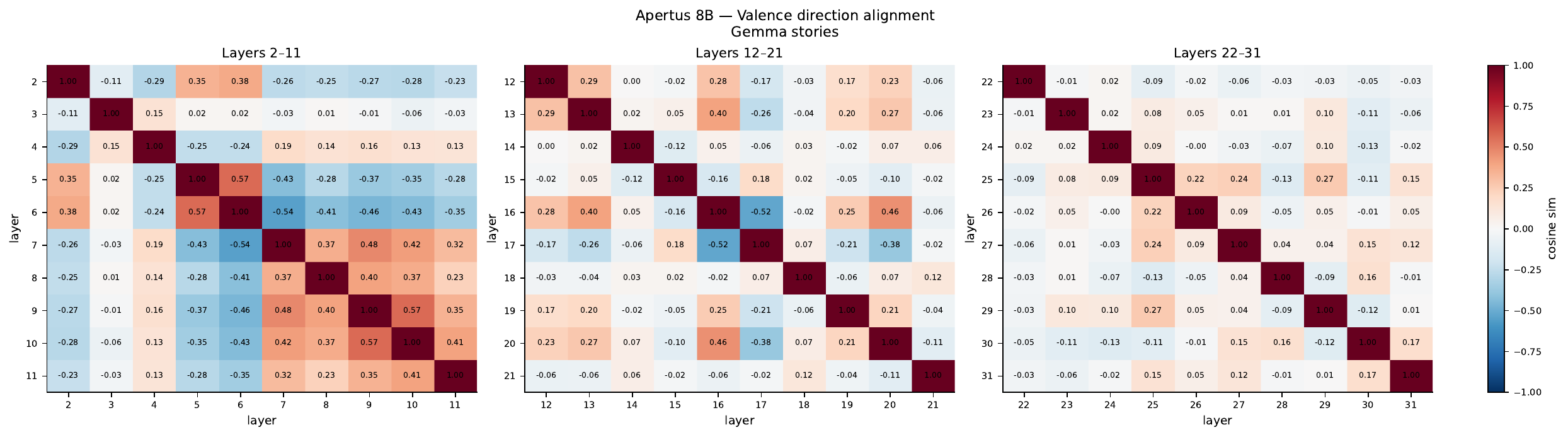}
\centering
 \label{fig:cka_apertus_gemma_panel_val}
\end{figure}

\begin{figure}[H]
\caption{\textbf{\apertus} validation on Apertus stories }
\includegraphics[width=1.0\textwidth]{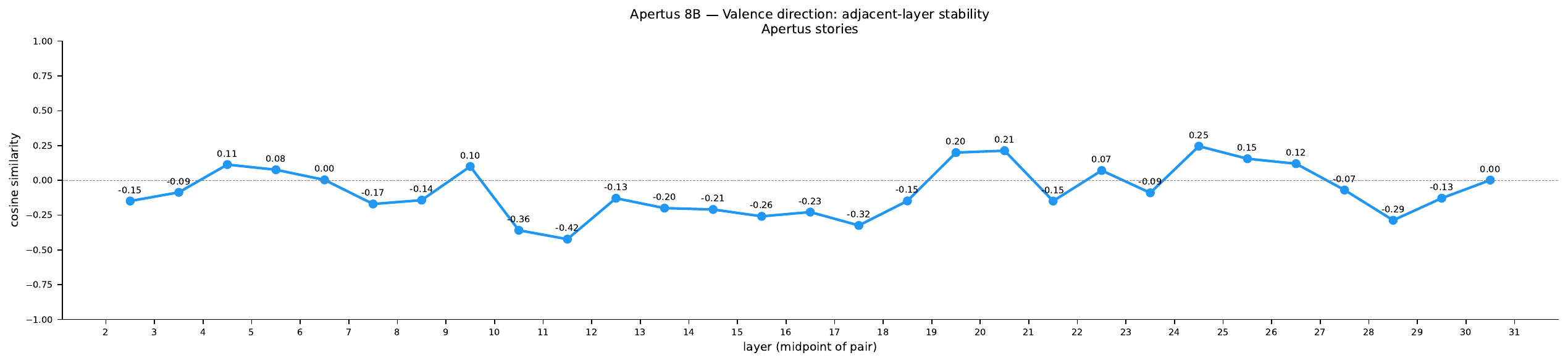}
\centering
 \label{fig:cka_apertus_apertus_val_stable}
\end{figure}

\begin{figure}[H]
\caption{\textbf{\apertus} validation on Gemma stories}
\includegraphics[width=1.0\textwidth]{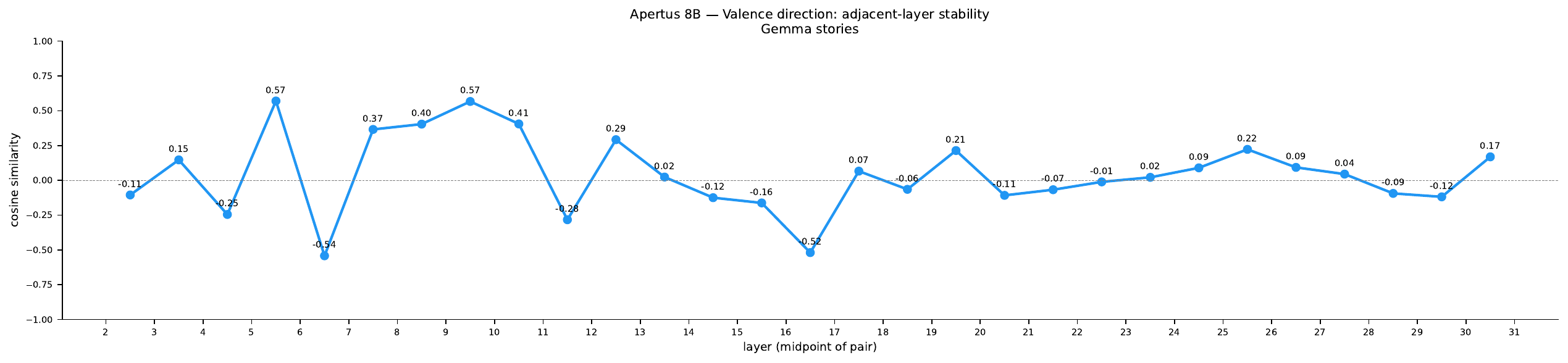}
\centering
 \label{fig:cka_apertus_gemma_val_stable}
\end{figure}

\subsubsection{\gemma Results}

\begin{figure}[H]
\caption{\textbf{\gemma} validation on Gemma stories }
\includegraphics[width=1.0\textwidth]{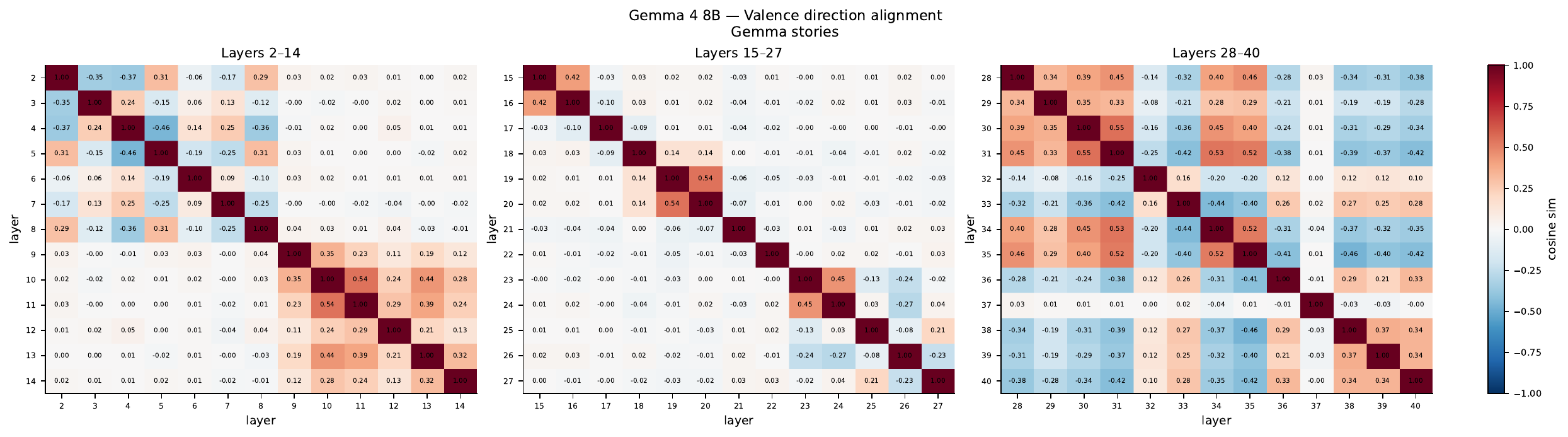}
\centering
 \label{fig:cka_gemma_gemma_panel_val}
\end{figure}

\begin{figure}[H]
\caption{\textbf{\gemma} validation on Apertus stories}
\includegraphics[width=1.0\textwidth]{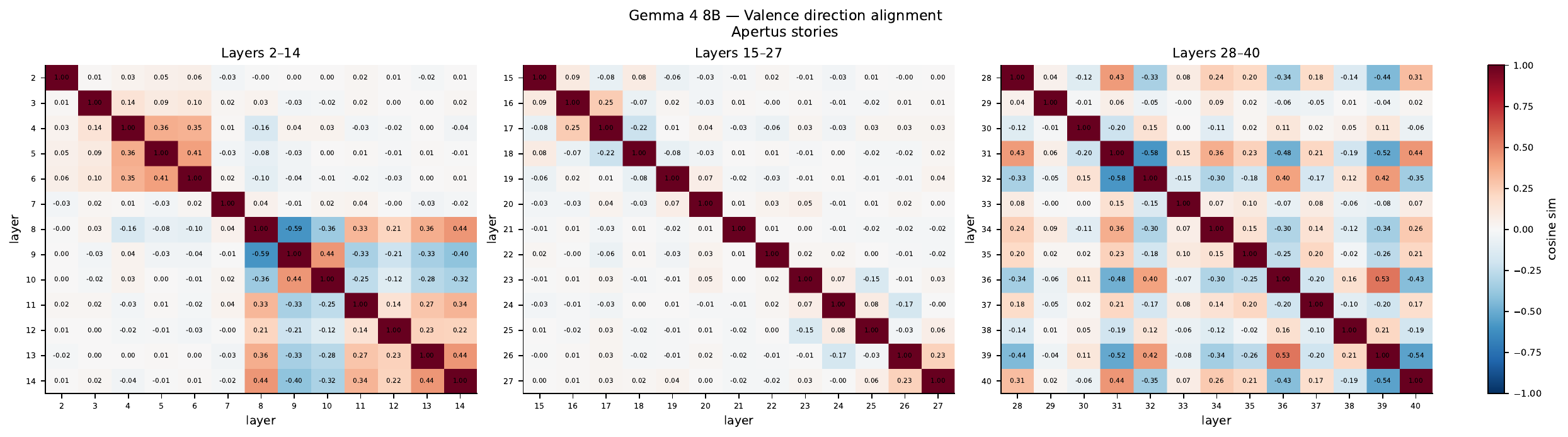}
\centering
 \label{fig:cka_gemma_apertus_panel_val}
\end{figure}

\begin{figure}[H]
\caption{\textbf{\gemma validation} on Apertus stories}
\includegraphics[width=1.0\textwidth]{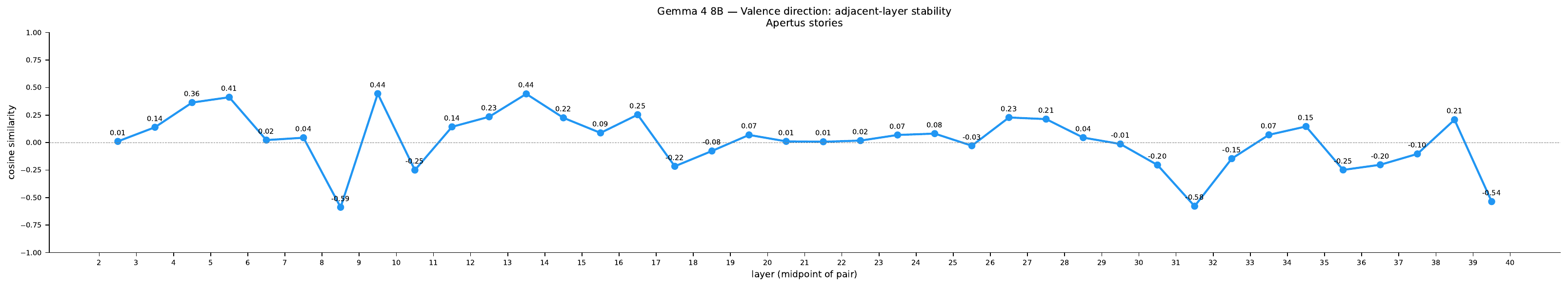}
\centering
 \label{fig:cka_gemma_apertus_val_stable}
\end{figure}

\begin{figure}[H]
\caption{\textbf{\gemma} on Gemma stories }
\includegraphics[width=1.0\textwidth]{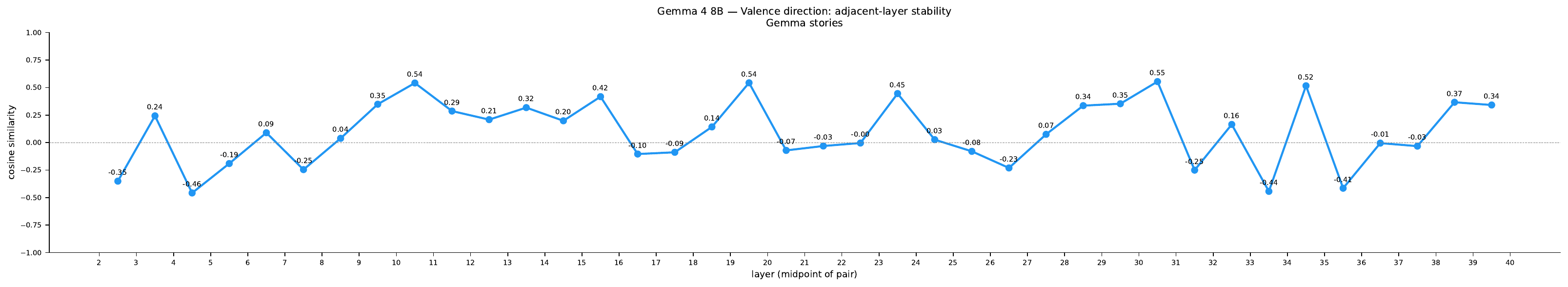}
\centering
 \label{fig:cka_gemma_gemma_val_stable}
\end{figure}

\subsection{PCA comparison}
The PCA figure shows a map of the model's emotional space at a specific layer. We pick the layer with the highest valence. Each dot is an emotion, positioned at how the model actually represents this emotion in its activation space. 

Comparing two panels can tell whether the map is reproducible across different inputs, or whether the emotional space is sensitive to what stories the model reads. 

\subsubsection{\apertus Results}

\begin{figure}[H]
\caption{\textbf{\apertus} validation }
\includegraphics[width=1.0\textwidth]{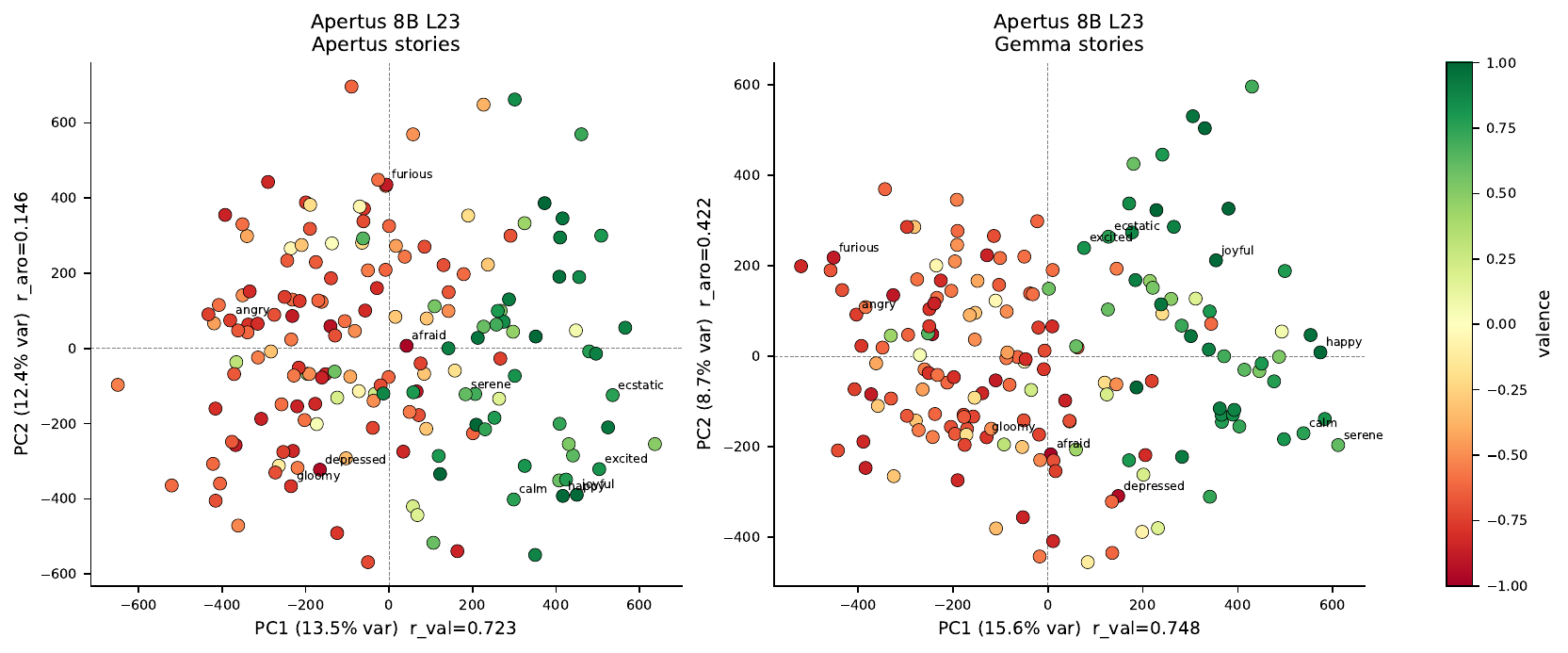}
\centering
 \label{fig:apertus_pca}
\end{figure}

\subsubsection{\gemma Results}

\begin{figure}[H]
\caption{\textbf{\gemma: valence-arousal PCA }}
\includegraphics[width=1.0\textwidth]{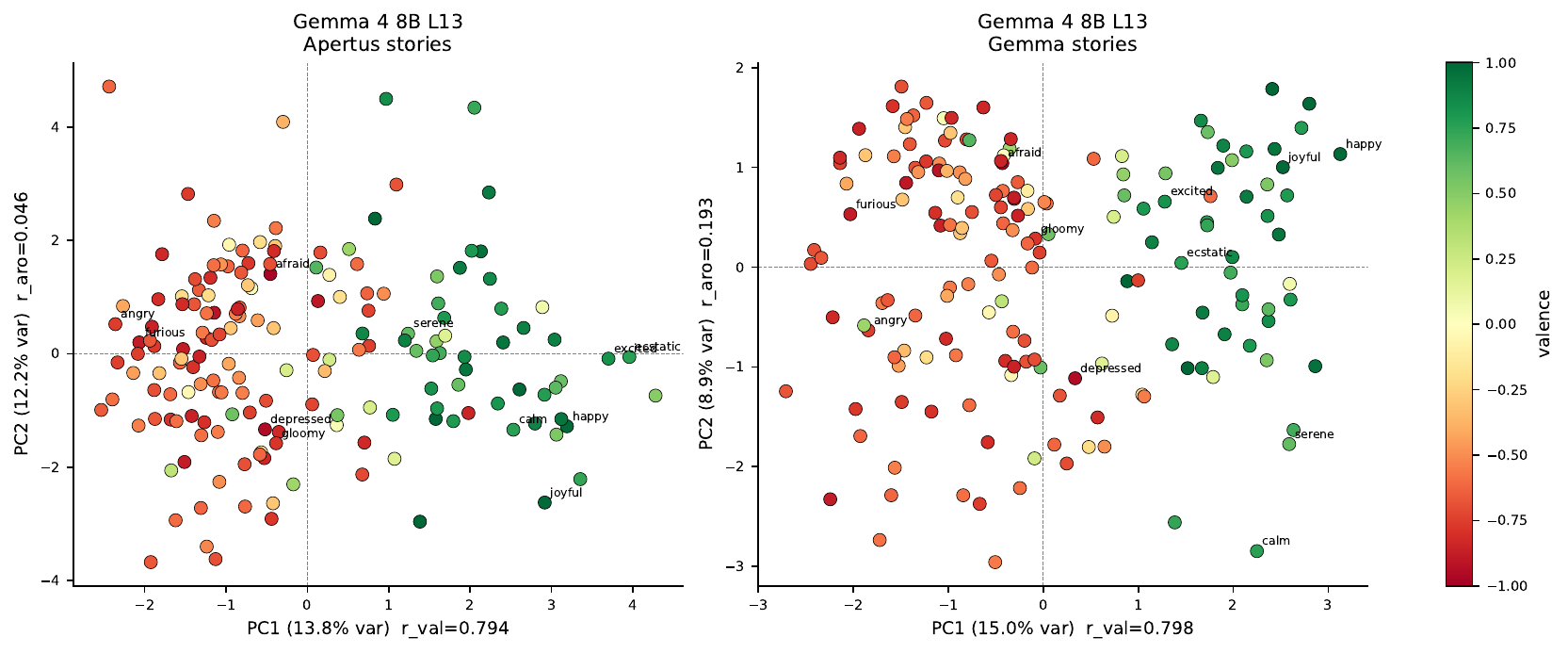}
\centering
 \label{fig:gemma_pca}
\end{figure}

\newpage
\section{Prompts}

Below, we report verbatim the prompts used to generate the short stories.

\begin{promptbox}{System prompt — Explanation generation}
Write {n_stories} different stories based on the following premise.
Topic: {topic}
The story should follow a character who is feeling {emotion}.
Format the stories like so:
[story 1]
[story 2]
[story 3]
etc.
The paragraphs should each be a fresh start, with no continuity. Try to make them diverse and not use the same turns of phrase. Across the different stories, use a mix of third-person narration and first-person narration.
IMPORTANT: You must NEVER use the word '{emotion}' or any direct synonyms of it in the stories. Instead, convey the emotion ONLY through:
- The character's actions and behaviors
- Physical sensations and body language
- Dialogue and tone of voice
- Thoughts and internal reactions
- Situational context and environmental descriptions
The emotion should be clearly conveyed to the reader through these indirect means, but never explicitly named.
\end{promptbox}









\end{document}